\documentclass[10pt,twocolumn]{article}
\usepackage[margin=0.75in]{geometry}
\usepackage[T1]{fontenc}
\usepackage{booktabs}
\usepackage{graphicx}
\usepackage{amsmath,amssymb}
\usepackage{microtype}
\usepackage[hidelinks]{hyperref}
\usepackage{xcolor}
\newcommand{\NAbstained}{14}
\newcommand{\NBgeBinCos}{$+$5.35}
\newcommand{\NBgeBinGap}{$+$2.89}
\newcommand{\NBgeBinRank}{$+$2.46}
\newcommand{\NBgeBins}{4}

\newcommand{\NBgeMaxDfunc}{0.201}

\newcommand{\NBgePairCIhi}{$+$2.38}
\newcommand{\NBgePairCIlo}{$-$5.18}

\newcommand{\NBgePairMaxdD}{0.008}
\newcommand{\NBgePairMean}{$-$1.40}
\newcommand{\NBgePairSD}{1.52}
\newcommand{\NBgeTol}{0.92}
\newcommand{\NBgeVerdict}{retrieval\_specific}
\newcommand{\NDecisions}{30}
\newcommand{\NDfuncEFiveMean}{0.032}
\newcommand{\NDfuncGemmaMean}{0.686}
\newcommand{\NDfuncRepeatAtBgeDose}{6e-3}
\newcommand{\NDfuncRepeatAtEFiveDose}{9e-4}
\newcommand{\NDfuncRepeatEFiveDoseD}{0.16}
\newcommand{\NDfuncRepeatMax}{2.7e-2}
\newcommand{\NDfuncRepeatMaxD}{0.39}
\newcommand{\NDfuncRepeatMaxStep}{10}
\newcommand{\NDfuncRepeatStepOne}{1e-8}
\newcommand{\NEFiveBinCos}{$-$9.83}
\newcommand{\NEFiveBinGap}{$-$13.05}
\newcommand{\NEFiveBinRank}{$+$3.22}
\newcommand{\NEFiveBins}{3}
\newcommand{\NEFiveCosDfunc}{0.148}
\newcommand{\NEFiveCosDrop}{$-$11.5}

\newcommand{\NEFiveLrA}{$+$2.07}

\newcommand{\NEFiveLrB}{$+$4.26}

\newcommand{\NEFiveLrC}{$+$6.05}

\newcommand{\NEFiveLrD}{$-$8.60}

\newcommand{\NEFivePairCIhi}{$+$24.96}
\newcommand{\NEFivePairCIlo}{$+$3.37}

\newcommand{\NEFivePairMaxdD}{0.016}
\newcommand{\NEFivePairMean}{$+$14.16}
\newcommand{\NEFivePairSD}{4.34}
\newcommand{\NEFiveTol}{0.58}
\newcommand{\NEFiveVerdict}{control\_worse}
\newcommand{\NGateOneNullMax}{0.7830}
\newcommand{\NGateOneNullN}{50}
\newcommand{\NGateOneP}{0.0196}
\newcommand{\NGateOneTreat}{1.0369}
\newcommand{\NGemmaBins}{7}
\newcommand{\NGemmaCollapseCos}{$-$69.52}
\newcommand{\NGemmaCollapseRank}{$-$71.63}

\newcommand{\NGemmaLrA}{$+$1.47}

\newcommand{\NGemmaLrB}{$-$6.28}

\newcommand{\NGemmaLrC}{$-$30.48}

\newcommand{\NGemmaLrD}{$-$71.53}

\newcommand{\NGemmaMseDfunc}{0.107}
\newcommand{\NGemmaNullExceed}{1}
\newcommand{\NGemmaNullMax}{$+$2.70}
\newcommand{\NGemmaNullN}{20}
\newcommand{\NGemmaTol}{1.51}
\newcommand{\NGemmaTreat}{$+$2.38}
\newcommand{\NGemmaVerdict}{not\_retrieval\_specific}
\newcommand{\NHarmPrereg}{0.56}
\newcommand{\NHarmful}{0}
\newcommand{\NLbAccessed}{1,024}
\newcommand{\NLbCIhi}{$+$0.038}
\newcommand{\NLbCIlo}{$-$0.037}
\newcommand{\NLbDiff}{$+$0.0004}

\newcommand{\NNRuns}{161}
\newcommand{\NOppAcc}{8}
\newcommand{\NOppN}{11}
\newcommand{\NParityEFive}{85,524,480}
\newcommand{\NParityGemma}{101,449,728}
\newcommand{\NParityLayersEFive}{73}
\newcommand{\NParityLayersGemma}{168}
\newcommand{\NPathResidual}{1.61}
\newcommand{\NRegretAcc}{0.49}
\newcommand{\NRegretCIhi}{2.76}
\newcommand{\NRegretCIlo}{0.59}
\newcommand{\NRegretMax}{11.13}
\newcommand{\NRegretMean}{1.52}
\newcommand{\NRegretSD}{2.66}
\newcommand{\NSigmaDfuncEFive}{0.00026}
\newcommand{\NSigmaDfuncEFiveRel}{0.8}
\newcommand{\NSigmaDfuncGemma}{0.0227}
\newcommand{\NSigmaDfuncGemmaRel}{3.3}
\newcommand{\NSigmaOptEFive}{0.192}
\newcommand{\NSigmaOptGemmaA}{0.504}
\newcommand{\NSigmaOptGemmaB}{1.141}
\newcommand{\NSigmaOptGemmaC}{3.470}
\newcommand{\NSigmaOptGemmaD}{0.675}
\newcommand{\NSigmaRuntimeEFive}{0.184}
\newcommand{\NSigmaSlateEFiveArg}{0.66}
\newcommand{\NSigmaSlateEFiveSci}{2.54}
\newcommand{\NSlates}{6}
\newcommand{\NTieArguana}{16}
\newcommand{\NTieScidocs}{10}
\newcommand{\NTieSwingArguana}{1.48}
\newcommand{\NTieSwingScidocs}{0.50}
\newcommand{\NZeroSixUpper}{39}

\newcommand{\Dfunc}{D_{\mathrm{func}}}
\newcommand{\pp}{\,\mathrm{pp}}

\date{2026-09-28}   
\title{When PTQ Calibration Helps or Hurts:\\ Dose-Driven Collapse and Validation-Gated Abstention\\ in Quantized Text Embedders}
\author{Hyojung Han$^{1,\dagger}$ \quad Ryangkyung Kang$^{1}$ \quad Youngeun Kim$^{2}$
\quad Seung-Hun Jeon$^{1}$\\[4pt]
\small $^{1}$ThakiCloud \qquad $^{2}$Korea University\\[2pt]
\small $^{\dagger}$Corresponding author: \texttt{hyojung.han@thakicloud.com}}

\begin{document}
\maketitle

\begin{abstract}
Learned post-training quantization (PTQ) calibration fits quantizer parameters on a small
calibration set, and for a text embedder the natural supervision is retrieval relevance. We
ask two questions that are usually conflated: \emph{how far} to calibrate, and \emph{whether}
to calibrate at all. Across three embedders (e5-base-v2, BGE-M3, EmbeddingGemma-300m) at 4-bit
weight quantization, \NNRuns{} calibration runs and five learning rates, we find that the outcome
of a fixed procedure ranges from a gain to catastrophic degradation, and that the driver is
the \emph{dose} of the update, measured as the functional displacement $\Dfunc$ of the quantized
model from its uncalibrated start. On EmbeddingGemma the same learning-rate grid gives
\NGemmaLrA{} to \NGemmaLrD{} points of in-domain NDCG@10 change; a label-free cosine
reconstruction objective collapses at the same dose (\NGemmaCollapseCos{} vs \NGemmaCollapseRank{}
in the highest dose bin), so the collapse is a property of learnable calibration rather than of
retrieval supervision. Objective sensitivity is nevertheless heterogeneous: on e5, ranking
calibration is substantially more robust than cosine calibration at matched displacement
(\NEFivePairMean{} $\pm$ \NEFivePairSD{}$\pp$, 95\% t-CI [\NEFivePairCIlo, \NEFivePairCIhi],
three paired slates), whereas on BGE-M3 the ordering is directionally opposite but unresolved
([\NBgePairCIlo, \NBgePairCIhi]). Dose is the main axis but not a sufficient statistic: equal
$\Dfunc$ reached along different paths differs by up to \NPathResidual$\pp$. We therefore treat
calibration as an optional intervention. A pre-registered rule that accepts a checkpoint only
when a per-query paired confidence interval on a 25\% calibration-validation split clears the
uncalibrated baseline was run on \NSlates{} independent calibration slates and five arms. It never
selected a checkpoint below its slate's baseline (\NHarmful{} harmful selections in \NDecisions{}
decisions; pre-registered harm-prevention score \NHarmPrereg), retained \NOppAcc{} of \NOppN{}
beneficial updates, and paid a mean oracle regret of \NRegretMean$\pp$ (slate-cluster 95\%
[\NRegretCIlo, \NRegretCIhi]) with one abstention that forfeited \NRegretMax$\pp$. The result is
a safety--opportunity trade-off, not a free improvement. We report what remains open: the ordering
on BGE-M3, the mechanism behind path residuals, and label efficiency, which our audit shows was
never established.
\end{abstract}

\section{Introduction}

Post-training quantization of a text embedder to 4-bit weights is cheap when the quantizer
parameters are set by a closed-form rule (min--max or percentile scaling) and expensive to get
wrong: the deployed artifact ranks documents, and a small shift in the embedding geometry can
reorder the top of a result list without changing any reconstruction statistic much. \emph{Learned}
calibration~\cite{esser2020lsq,choi2018pact,nagel2020adaround,li2021brecq,shao2024omniquant} fits those parameters with gradient
steps on a small calibration set, and for a retriever the tempting supervision is retrieval
relevance itself---a ranking loss over labelled queries~\cite{burges2007lambdarank}.

This paper is about what that procedure actually does to three embedders, measured carefully
enough to separate two questions that are usually asked together. The first is a \emph{dose}
question: how far should the quantizer move from its uncalibrated start? The second is a
\emph{gate} question: should it move at all, and can that be decided without touching held-out
labels? We find that the first question has no universal answer, that the second one has a
usable partial answer, and that several claims one would naturally make along the way---that
retrieval supervision causes the collapse, that a ranking objective is safer, that selected
calibration queries are label-efficient---are not supported by the ledger once the comparison
is dose-matched, slate-paired, or counted in the right unit.

\paragraph{Contributions.} (1) On EmbeddingGemma-300m the outcome of one fixed procedure spans
\NGemmaLrA{} to \NGemmaLrD$\pp$ of in-domain NDCG@10 across a four-point learning-rate grid, and
the collapse is reproduced by a label-free objective at the same functional displacement; the
unsafe regime belongs to learnable calibration, not to retrieval supervision (\S\ref{sec:dose}).
(2) Objective sensitivity is heterogeneous across models: a large ranking-over-cosine robustness
advantage on e5, no established ordering on BGE-M3 (\S\ref{sec:objective}). (3) $\Dfunc$ is a
useful screening variable but not a sufficient statistic; a pre-registered scalar-threshold
hypothesis is rejected (\S\ref{sec:distance}). (4) A pre-registered validation rule, run on six
independent calibration slates, never selected a checkpoint below the uncalibrated baseline while
forfeiting some beneficial updates; we report it as a four-number scorecard, keeping the
pre-registered metric beside the realized outcome (\S\ref{sec:gating}). (5) An implementation
audit that changed several conclusions is reported as part of the result, including a label-budget
accounting that retracts an earlier efficiency claim (\S\ref{sec:validity}, \S\ref{sec:notclaimed}).

We claim model-agnosticism of the \emph{procedure}, not model-invariance of the \emph{numbers}.
The three embedders differ in backbone, pooling, prefix conventions and normalization; a single
safe-dose constant is neither expected nor observed. What generalizes is that one frozen rule
adapts to each model through its own validation signal---model A receiving a small update and
model B receiving none is the content of the generalization, not a failure of it.

\section{Experimental setup}
\label{sec:setup}

\paragraph{Models.} e5-base-v2~\cite{wang2022e5} (BERT~\cite{devlin2019bert} backbone, mean pooling, ``query:/passage:''
prefixes), BGE-M3~\cite{chen2024bgem3} (XLM-R~\cite{conneau2020xlmr} backbone, CLS pooling, no prefix) and
EmbeddingGemma-300m~\cite{vera2025embeddinggemma} (Gemma~3~\cite{gemma2025gemma3} decoder backbone, mean pooling with a dense head,
task prefixes). All linear layers are quantized to 4-bit weights with a learned step-size quantizer
(LSQ~\cite{esser2020lsq}, with rounding passed through the straight-through
estimator~\cite{bengio2013ste}); the quantizer parameters are the only trainable variables.

\paragraph{Data.} Calibration draws 32 labelled queries from each of four BEIR~\cite{thakur2021beir}
domains (MS~MARCO~\cite{nguyen2016msmarco}, FiQA, NFCorpus, SciFact~\cite{wadden2020scifact}) over a 2,000-document corpus per
domain. In-domain generalization is measured on 25 \emph{disjoint} queries per domain that share no
query with the calibration set. Cross-domain generalization is measured on two held-out datasets
never used for calibration (ArguAna, SciDocs~\cite{cohan2020specter}; 100 queries, 2,000 documents). Three further datasets
(TREC-COVID~\cite{voorhees2020treccovid}, Climate-FEVER~\cite{diggelmann2020climatefever}, Quora) remain sealed and were not evaluated in any run reported here.
MS~MARCO and SciFact are ceiling-limited in this setting (more than 70\% of full-precision queries
already at NDCG@10\,=\,1.0), so we keep the four-domain macro average only as a continuity figure
and read improvement on the two unsaturated domains and on the held-out sets.

\paragraph{Objectives.} The retrieval-supervised objective is LambdaRank~\cite{burges2007lambdarank}
over each query's positives and mined negatives (\texttt{lambdarank}). Two label-free controls
share every other setting: \texttt{cosine}, which pulls quantized embeddings toward their
full-precision counterparts, and \texttt{mse}, which reconstructs weights. Calibration runs for
12 steps with Adam~\cite{kingma2015adam} at a learning rate from $\{10^{-4}, 3{\times}10^{-4}, 10^{-3}, 3{\times}10^{-3}\}$,
extended to $10^{-2}$ where a control could not otherwise reach the dose of interest.

\paragraph{Dose.} $\Dfunc$ is the mean cosine distance between the calibrated and the
uncalibrated quantized embeddings of the calibration queries and their candidate documents at a
given step. It measures how far the quantized \emph{function} has moved, independently of the
objective that moved it, and is the matching variable for every cross-objective comparison.

\paragraph{Metrics and noise.} We report NDCG@10 change in percentage points (pp) relative to
the uncalibrated quantized model (step~0) unless stated otherwise; retention is the quantized/full-
precision ratio. Noise floors are measured per model and per learning rate rather than borrowed:
on e5, re-running the same configuration gives $\sigma_{\text{runtime}}=\NSigmaRuntimeEFive\pp$
and varying only the optimizer seed $\sigma_{\text{opt}}=\NSigmaOptEFive\pp$ (SciDocs), while varying only
the calibration slate gives $\sigma_{\text{slate}}=\NSigmaSlateEFiveArg\pp$ on ArguAna and \NSigmaSlateEFiveSci$\pp$ on
SciDocs---the slate, not the optimizer, is the dominant variance source. On EmbeddingGemma $\sigma_{\text{opt}}$
is heteroscedastic across the learning-rate grid (\NSigmaOptGemmaA{}, \NSigmaOptGemmaB{},
\NSigmaOptGemmaC{}, \NSigmaOptGemmaD$\pp$), so significance is judged against the cell's own
$\sigma$ and repeated statistics are reported as median [min, max] or mean $\pm$ SD with $n$ stated.
Every headline comparison below is paired within a calibration slate.

\paragraph{Pre-registration.} The selection protocol (\S\ref{sec:gating}), the dose-matched
comparator (\S\ref{sec:dose}), the saturation criterion and the hypotheses in
\S\ref{sec:notclaimed} were frozen before the corresponding batches ran; where a post-hoc analysis
is added it is labelled as such. All numbers in this paper are generated from the experiment ledger
by code; none is typed by hand.

\section{Result 1: collapse is a dose effect, and not retrieval-specific}
\label{sec:dose}

\begin{figure*}[t]
\centering
\includegraphics[width=\textwidth]{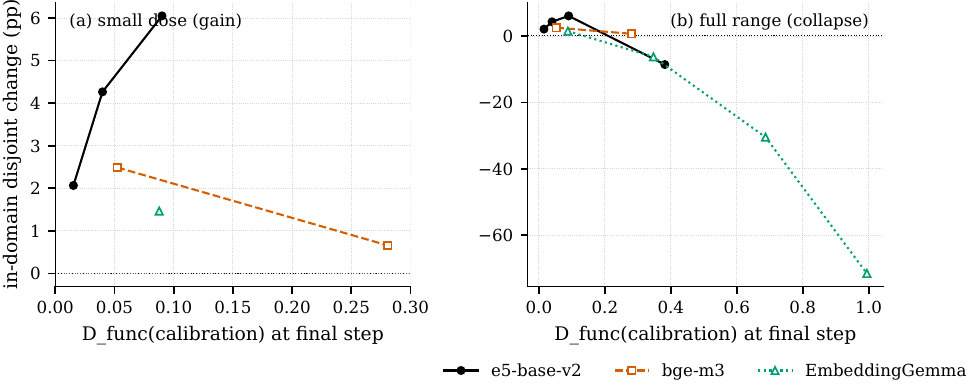}
\caption{In-domain disjoint NDCG@10 change (pp, final step vs step~0) against functional
displacement $\Dfunc$ at the final step, one point per learning rate (median of three optimizer
seeds), LambdaRank objective. (a) the small-dose region where every model gains; (b) the full
range. $\Dfunc$ is not comparable across models---each model is its own series---but within a
model the damage is monotone in dose. EmbeddingGemma reaches $\Dfunc\approx1$ at $3{\times}10^{-3}$
and loses \NGemmaLrD$\pp$.}
\label{fig:dose}
\end{figure*}

On EmbeddingGemma the four learning rates give \NGemmaLrA{}, \NGemmaLrB{}, \NGemmaLrC{} and
\NGemmaLrD$\pp$ (medians of three optimizer seeds; Fig.~\ref{fig:dose}). The same procedure that
helps at $10^{-4}$ destroys the model at $3{\times}10^{-3}$. e5 shows the same shape at a
different scale (\NEFiveLrA{}, \NEFiveLrB{}, \NEFiveLrC{}, \NEFiveLrD$\pp$): an earlier draft of
this work read the two models as an architectural dichotomy---``works on e5, fails on Gemma''---and
that reading was an uncontrolled learning-rate comparison, not a property of either architecture.

\paragraph{Is the collapse caused by retrieval supervision?} We ran the label-free controls on
the same slates, quantizer parameters, learning-rate grid, steps and evaluation queries, changing
only the objective. Read by learning rate the generic objectives look stable, but at the same
learning rate they move the quantizer four to six times less; the pre-registered comparison is
therefore \emph{dose-matched}: points from all trajectories are binned by $\Dfunc$ and the two
objectives are compared within each bin against a tolerance of three times the model's own
$\sigma_{\text{opt}}$ (Table~\ref{tab:dose}). On EmbeddingGemma the cosine objective collapses
in the highest dose bin exactly as LambdaRank does (\NGemmaCollapseCos{} vs
\NGemmaCollapseRank$\pp$, both saturated), and after filling the intermediate doses with two extra
learning rates ($1.5{\times}10^{-3}$, $2{\times}10^{-3}$, $n{=}3$ each) the comparator has
\NGemmaBins{} qualifying bins and returns \texttt{\NGemmaVerdict}: at every matched dose the
generic objective is hurt at least as much as the ranking one. The weight-reconstruction control
never reaches the collapse dose---its loss converges quickly and the quantizer stops at
$\Dfunc=\NGemmaMseDfunc$ even at the largest learning rate---which is not evidence of safety but
of an objective that cannot deliver the dose.

\begin{table}[t]
\centering\small
\caption{Pre-registered dose-matched comparison of the ranking objective against the label-free
cosine objective, per model and $\Dfunc$ bin: mean in-domain disjoint change (pp) of all trajectory
points falling in the bin, and the gap (ranking $-$ cosine). Tolerance is three times each model's
own $\sigma_{\text{opt}}$; the verdict is the comparator's, computed on slate~0 trajectories.
EmbeddingGemma bins 0.20--0.75 come from the added intermediate learning rates. Verdicts: Gemma \texttt{\NGemmaVerdict} (\NGemmaBins{} bins, tolerance \NGemmaTol$\pp$), e5 \texttt{\NEFiveVerdict} (\NEFiveBins{} bins, \NEFiveTol$\pp$), BGE-M3 \texttt{\NBgeVerdict} (\NBgeBins{} bins, \NBgeTol$\pp$).}
\label{tab:dose}
\footnotesize\begin{tabular}{@{}llrrr@{}}\toprule
Model & $D_{\mathrm{func}}$ bin & ranking & cosine & gap \\ \midrule
Gemma & 0.00--0.05 & $+$0.88 & $+$1.19 & $+$0.31 \\
 & 0.05--0.10 & $+$1.27 & $+$1.08 & $-$0.19 \\
 & 0.10--0.20 & $+$0.09 & $-$3.97 & $-$4.07 \\
 & 0.20--0.35 & $-$3.53 & $-$11.77 & $-$8.24 \\
 & 0.35--0.50 & $-$12.18 & $-$13.55 & $-$1.37 \\
 & 0.50--0.75 & $-$26.09 & $-$23.62 & $+$2.48 \\
 & 0.75--1.01 & $-$71.63 & $-$69.52 & $+$2.11 \\
e5 & 0.00--0.05 & $+$2.36 & $+$3.23 & $+$0.88 \\
 & 0.05--0.10 & $+$5.60 & $+$5.23 & $-$0.37 \\
 & 0.10--0.20 & $+$3.22 & $-$9.83 & $-$13.05 \\
BGE-M3 & 0.00--0.05 & $+$1.85 & $+$1.92 & $+$0.07 \\
 & 0.05--0.10 & $+$2.24 & $+$3.22 & $+$0.97 \\
 & 0.10--0.20 & $+$3.91 & $+$5.36 & $+$1.45 \\
 & 0.20--0.35 & $+$2.46 & $+$5.35 & $+$2.89 \\
\bottomrule\end{tabular}

\end{table}

The existence of an unsafe regime is thus a property of learnable calibration under a large
enough update, not of reading relevance labels. This is the claim the rest of the paper is built
on: \emph{PTQ calibration is an intervention with a dose, and the dose can be lethal.}

\section{Result 2: objective sensitivity is heterogeneous across models}
\label{sec:objective}

Whether the two objectives differ \emph{within} the safe regime is a separate question, and the
answer differs by model. On e5 the cosine objective stays near $\Dfunc\approx0.06$ across the
original grid and only reaches the collapse dose at $10^{-2}$ ($\Dfunc=\NEFiveCosDfunc$,
\NEFiveCosDrop$\pp$, $n{=}3$); in the same $\Dfunc$ bin the ranking objective is still positive
(\NEFiveBinRank{} vs \NEFiveBinCos$\pp$, gap \NEFiveBinGap), and the comparator returns
\texttt{\NEFiveVerdict}. On BGE-M3 neither objective collapses anywhere on the grid
(cosine reaches at most $\Dfunc=\NBgeMaxDfunc$), and the slate-0 comparator favours the cosine
objective (\NBgeBinCos{} vs \NBgeBinRank$\pp$ in the 0.20--0.35 bin, gap \NBgeBinGap{} against a
tolerance of \NBgeTol$\pp$; \texttt{\NBgeVerdict}).

\begin{table*}[t]
\centering\small
\caption{Paired-slate, step-matched objective comparison. For each of three independent
calibration slates the two decisive cells per model are run; on the higher-dose arm the measured
step whose $\Dfunc$ is nearest the lower arm's final $\Dfunc$ is selected (caliper 0.02, met by
every pair), and $\Delta$ is the ranking-minus-cosine in-domain disjoint change at that matched
dose. $\pm$ is the SD over slates; the interval is a 95\% $t$-CI with $n{=}3$. ``interp.'' repeats
the analysis with linear interpolation between adjacent measured steps and is a sensitivity check,
not an observation.}
\label{tab:pairs}
\footnotesize\begin{tabular}{@{}llrrrrr@{}}\toprule
Model & cells & $\Delta$ per slate & mean $\pm$ SD & 95\% t-CI & interp.\ & max $|\Delta D|$ \\ \midrule
e5 & rank 3e-3 / cos 1e-2 ($D{\approx}0.15$) & $+$16.98 / $+$9.16 / $+$16.35 & $+$14.16 $\pm$ 4.34 & [$+$3.37, $+$24.96] & $+$14.08 & 0.016 \\
BGE-M3 & rank 1e-3 / cos 3e-3 ($D{\approx}0.21$) & $-$3.14 / $-$0.70 / $-$0.35 & $-$1.40 $\pm$ 1.52 & [$-$5.18, $+$2.38] & $-$1.53 & 0.008 \\
\bottomrule\end{tabular}

\end{table*}

A single-slate bin comparison is the weaker instrument, so the two decisive cells per model were
re-run on three independent slates and matched step by step (Table~\ref{tab:pairs}). On e5 the
ranking advantage is large and clear of zero: $\Delta=\NEFivePairMean\pm\NEFivePairSD\pp$,
95\% t-CI [\NEFivePairCIlo, \NEFivePairCIhi]. On BGE-M3 the direction is the opposite but the
interval contains zero widely: $\NBgePairMean\pm\NBgePairSD\pp$, [\NBgePairCIlo, \NBgePairCIhi].
We place the paired multi-slate result above the single-slate comparator and state the outcome
accordingly: on e5, ranking calibration is substantially more robust than cosine calibration at
matched functional displacement; on BGE-M3 the ordering is unresolved; on EmbeddingGemma both
collapse. We do not claim that the ordering reverses by model, only that objective sensitivity
is heterogeneous, and we do not claim that either objective is generally safer.

\section{Result 3: distance explains damage but is not sufficient}
\label{sec:distance}

$\Dfunc$ is a far better explanatory variable than the step count: within a model the damage is
monotone in $\Dfunc$ (Fig.~\ref{fig:dose}), and it is the variable that makes the objective
comparison of \S\ref{sec:dose} possible at all. The pre-registered strong form---that performance
is a function of $\Dfunc$ alone---is nevertheless rejected: on e5, trajectories reaching the same
$\Dfunc$ band along different learning rates differ by up to \NPathResidual$\pp$ in
calibration-validation NDCG@10 (median per learning rate within the band), roughly nine times
$\sigma_{\text{runtime}}$. The measurement itself is not the reason. Repeating the collapsing
EmbeddingGemma configuration over five optimizer seeds gives $\Dfunc=\NDfuncGemmaMean$ with
$\sigma=\NSigmaDfuncGemma$ (\NSigmaDfuncGemmaRel\%), and at low dose on e5 five repeats give $\Dfunc=\NDfuncEFiveMean$ with $\sigma=\NSigmaDfuncEFive$
(\NSigmaDfuncEFiveRel\%); the path residual is real. (Whether the higher learning rate reaches the
band in fewer steps is confounded with how many updates the objective received, so we do not
attribute the residual to step size.)\footnote{Repeatability of $\Dfunc$ itself was measured by running one e5 configuration
(lr $3{\times}10^{-3}$, same slate and optimizer seed) twice on CPU: the trajectories agree to
\NDfuncRepeatStepOne{} at step~1 and diverge as floating-point non-determinism compounds through the
optimizer---\NDfuncRepeatAtEFiveDose{} at $\Dfunc\approx\NDfuncRepeatEFiveDoseD$, up to \NDfuncRepeatAtBgeDose{} for
$\Dfunc\in[0.20,0.33]$ and \NDfuncRepeatMax{} by step \NDfuncRepeatMaxStep{} ($\Dfunc\approx\NDfuncRepeatMaxD$). The matched
pairs of Table~\ref{tab:pairs} sit at $\Dfunc\approx0.15$ (e5, $|\Delta D|\le\NEFivePairMaxdD$) and $\approx0.21$
(BGE-M3, $|\Delta D|\le\NBgePairMaxdD$), roughly ten and two to four times the floor at those doses; the 0.02
caliper is above the instrument at both, but the BGE-M3 margin is thin, which is one more reason to
read that comparison as unresolved.} The honest statement is that the outcome is a
function of dose, objective, model and optimization path, with dose as the main axis. We do not
offer a mechanism for the residual and did not search for one; \S\ref{sec:notclaimed} lists it as
open.

\section{Result 4: selective calibration}
\label{sec:gating}

If calibration can help or hurt, the operational question is whether the harmful case can be
refused without held-out labels. The rule is simple and was frozen before the multi-slate batch:
hold out 25\% of the calibration queries; after each step, compute the per-query NDCG@10 gain over
step~0 on that split; accept the step whose paired bootstrap lower bound clears zero with the
largest mean gain; if no step clears, abstain and ship the uncalibrated quantizer. The held-out
sets never enter the rule. An earlier version of this section reported a plain argmax over the
validation mean; on three optimizer seeds that argmax chose a checkpoint below the baseline once
($-3.42\pp$ against oracles of about $+1.7\pp$), which is why the confidence-interval rule replaced it.

\begin{figure*}[t]
\centering
\includegraphics[width=\textwidth]{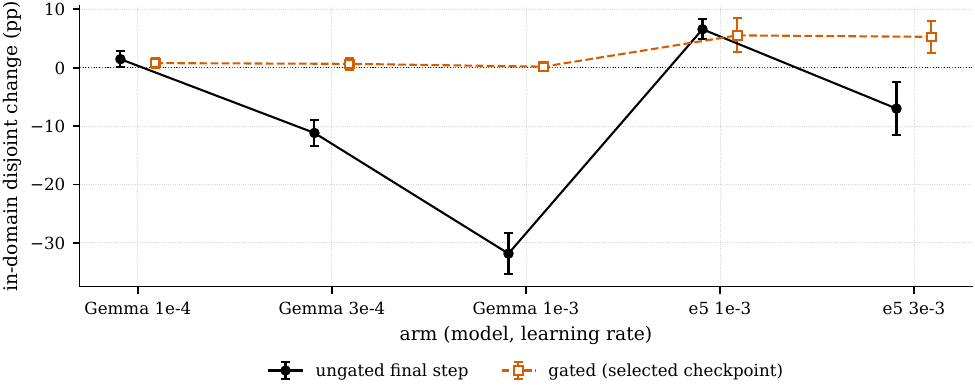}
\caption{Validation-gated selection on six independent calibration slates. For each arm (model,
learning rate) the filled series is the in-domain disjoint change at the final step without a
gate and the open series is the change at the checkpoint the rule selected (0 when it abstained);
error bars are SD over the six slates. The gated series never goes below zero; the price is
visible on the e5 arms, where the gate sometimes abstains from a large gain.}
\label{fig:gating}
\end{figure*}

\begin{table*}[t]
\centering\small
\caption{Six-slate scorecard, five arms. ``ungated final'' is the change at step~12 without a
gate, ``gated $\Delta$'' the change at the selected checkpoint (abstention scores 0), both mean
$\pm$ SD over six slates; ``min gated'' is the worst selected checkpoint on any slate; ``regret''
is the mean oracle regret over all six decisions of the arm, with abstention scored at the
baseline. The pre-registered harm-prevention score counts only abstention on arms whose
\emph{final} step is catastrophic and is kept unchanged beside the realized outcome.}
\label{tab:gating}
\footnotesize\begin{tabular}{@{}llrrrrr@{}}\toprule
Model & lr & ungated final & gated $\Delta$ & min gated & abstain & regret \\ \midrule
EmbeddingGemma & 1e-4 & $+$1.5 $\pm$ 1.4 & $+$0.78 $\pm$ 0.90 & $+$0.00 & 3/6 & 1.12 \\
EmbeddingGemma & 3e-4 & $-$11.2 $\pm$ 2.2 & $+$0.66 $\pm$ 1.05 & $+$0.00 & 4/6 & 1.33 \\
EmbeddingGemma & 1e-3 & $-$31.9 $\pm$ 3.5 & $+$0.18 $\pm$ 0.43 & $+$0.00 & 5/6 & 0.63 \\
e5-base-v2 & 1e-3 & $+$6.6 $\pm$ 1.7 & $+$5.53 $\pm$ 2.92 & $+$0.00 & 1/6 & 2.33 \\
e5-base-v2 & 3e-3 & $-$7.0 $\pm$ 4.5 & $+$5.29 $\pm$ 2.73 & $+$0.00 & 1/6 & 2.17 \\
\midrule
\multicolumn{7}{@{}l}{all 30 decisions: harmful selections 0 $\cdot$ abstained 14 $\cdot$ opportunity 8/11} \\
\multicolumn{7}{@{}l}{regret 1.52\,pp (max 11.13) $\cdot$ pre-registered harm\_prevention 0.56} \\
\bottomrule\end{tabular}

\end{table*}

The rule was run on five arms---EmbeddingGemma at $10^{-4}$, $3{\times}10^{-4}$, $10^{-3}$ and e5 at
$10^{-3}$, $3{\times}10^{-3}$---on six independent calibration slates (Table~\ref{tab:gating},
Fig.~\ref{fig:gating}). Across six independent slates and five arms per slate, no selected checkpoint
fell below its slate's baseline: \NHarmful{} harmful selections in \NDecisions{} decisions. The five
arms within a slate are correlated, so we do not treat this as thirty independent trials; taking
the slate as the unit, zero events in six gives a one-sided 95\% upper bound of about
\NZeroSixUpper\% on the per-slate probability of a harmful selection. The statement that holds is
that no harmful checkpoint was observed, not that the probability is near zero. The pre-registered
metric is reported unchanged: \texttt{harm\_prevention}\,$=\NHarmPrereg$, because it scores
abstention on arms whose final step collapses and the rule increasingly stopped \emph{early}
instead---on the EmbeddingGemma $3{\times}10^{-4}$ and $10^{-3}$ arms it accepted checkpoints of
$+1$ to $+2\pp$ on slates where the final step was $-10$ to $-32\pp$. We keep both numbers because
they answer different questions; a confirmation study would pre-register the realized harmful
selection rate as the primary safety endpoint.

The cost is opportunity. Regret is computed over all \NDecisions{} decisions with abstention scored
at the baseline: mean \NRegretMean$\pp$ (SD \NRegretSD; slate-cluster bootstrap 95\%
[\NRegretCIlo, \NRegretCIhi]), abstention rate \NAbstained/\NDecisions, opportunity retention
\NOppAcc/\NOppN, and a maximum regret of \NRegretMax$\pp$ on one e5 slate where the rule abstained
from both arms although the oracle checkpoint was strongly positive. Regret over accepted runs
alone is \NRegretAcc$\pp$ and is reported only as an auxiliary figure, since it hides exactly the
forfeited opportunities that matter. The selector is therefore conservative with respect to
catastrophic harm and sometimes conservative with respect to large gains; it is a safety--opportunity
trade-off, not a free improvement.

Two further observations bound what the small EmbeddingGemma gains mean. Doubling the validation
split to 50\% did not recover the missed $+2\pp$ updates (still one of three slates accepted) and
only shrank the optimization set. And the pre-registered shuffled-label null run on EmbeddingGemma
(\NGemmaNullN{} seeds, same cell) passes on the pre-registered metrics but the in-domain gain
of \NGemmaTreat$\pp$ sits inside the shuffled-label tail (null maximum \NGemmaNullMax$\pp$,
\NGemmaNullExceed{} exceedance): the small gain is scored as a missed opportunity in the decision
layer and is not attributed to retrieval supervision in the mechanism layer.

\section{Validity}
\label{sec:validity}

\paragraph{Fake-quantization.} All numbers come from simulated quantization, so every calibrated
checkpoint is also exported to packed INT4 integer codes and compared bit by bit against the
simulated path: \NParityEFive{} weights across \NParityLayersEFive{} layers on e5 and
\NParityGemma{} across \NParityLayersGemma{} on EmbeddingGemma, zero mismatches, at exactly four bits
per weight. The check runs by default on every run, and objectives without an integer code (soft
rounding) are refused export.

\paragraph{Evaluation determinism.} Retrieval scores are float32 on GPU and ties near the
cut-off can flip NDCG@10 between otherwise identical runs. An audit records, per held-out
query, the distance from the best positive to the nearest irrelevant score; over all slate runs the worst case on ArguAna is
\NTieArguana{} near-ties below $10^{-4}$ with a tie swing of \NTieSwingArguana$\pp$, and on SciDocs
\NTieScidocs{} near-ties with a swing of \NTieSwingScidocs$\pp$; differences below those swings are
below the instrument and are not headlined.

\paragraph{Label specificity.} On e5 the LambdaRank gain exceeds all 50 within-domain
shuffled-label controls (calibration-fit retention \NGateOneTreat{} vs a null maximum of
\NGateOneNullMax{} over \NGateOneNullN{} seeds; randomization $p=\NGateOneP$), and the same holds for the pre-registered metrics on
EmbeddingGemma. A derangement null and a same-code re-run of the plain null were added as
sensitivity analyses; neither changes the verdict.

\paragraph{Audit.} A two-pass implementation audit changed several conclusions before this
version and is part of the record: a loss misclassified as label-free, a query-selection routine
that read the relevance labels of its whole candidate pool, an under-exclusion of self-matches in
held-out corpora, a determinism audit that first measured the wrong gap, and a number gate whose
discriminating power turned out to be 54\% on perturbed values and was demoted to an existence
check in favour of row-level tests against the ledger.

\section{What we do not claim}
\label{sec:notclaimed}

\begin{table}[t]
\centering\small
\caption{Label-budget comparison (PRD gate ``128 selected $\approx$ 512 random, difference
$\le0.2$ NDCG''), in the gate's own unit. Held-out NDCG@10 is the mean over ArguAna and SciDocs;
``labels read'' is the number of relevance judgements the run accessed, including the candidate
pool screened by the selection heuristic.}
\label{tab:lb}
\footnotesize\begin{tabular}{@{}lrrr@{}}\toprule
calibration set & $n$ & held-out NDCG@10 & labels read \\ \midrule
128 random & 3 & 0.4755 $\pm$ 0.0350 & 128 \\
128 selected & 3 & 0.4835 $\pm$ 0.0143 & 1,024 \\
512 random & 3 & 0.4831 $\pm$ 0.0183 & 512 \\
512 selected & 3 & 0.4804 $\pm$ 0.0200 & 1,024 \\
\midrule
\multicolumn{4}{@{}l}{128 sel.\ $-$ 512 rnd.: $+$0.0004 NDCG} \\
\multicolumn{4}{@{}l}{Welch 95\% CI [$-$0.037, $+$0.038]} \\
\bottomrule\end{tabular}

\end{table}

\paragraph{Label efficiency.} An earlier draft described the selected calibration set as
``32 labelled queries''. The audit showed that the margin-diversity selector reads the relevance
labels of its entire candidate pool, so the labels \emph{accessed} exceed the labels
\emph{selected}, which exceed the labels the gradient \emph{uses}. In the gate's own unit
(Table~\ref{tab:lb}) 128 selected and 512 random calibration examples give indistinguishable
point estimates at the present sample size (\NLbDiff{} NDCG, Welch 95\% CI [\NLbCIlo, \NLbCIhi],
$n{=}3$ each), while the selected arm accessed \NLbAccessed{} labels against 512. The pre-registered
margin of 0.2 NDCG is wider than any plausible effect and is recorded as a specification defect:
the gate passed without information. We therefore make no label-efficiency claim; the supportable
observation is that a smaller selected calibration set produced a similar point estimate to a
four-times-larger random one, at higher label access.

\paragraph{The BGE-M3 ordering} is unresolved (\S\ref{sec:objective}), and BGE-M3 has not been
pushed to a collapsing dose; ``every model collapses if pushed far enough'' is not a claim this
paper supports.

\paragraph{Mechanism.} The path residual of \S\ref{sec:distance} and the model dependence of
objective sensitivity have no mechanistic account here; update direction, objective geometry,
representation geometry, curvature and quantizer parameterization are all candidates, and we did
not begin that search because it would have reopened dozens of variables before the descriptive
results were pinned.

\paragraph{Scale and sealed data.} Held-out evaluation uses 100 queries over 2,000 documents per
dataset, so absolute NDCG is not headlined; three test datasets remain sealed for a single
post-freeze evaluation that this version has not performed.

\section{Related work}

Post-training quantization is surveyed in~\cite{gholami2021survey}. Learned PTQ
calibration~\cite{esser2020lsq,choi2018pact,nagel2020adaround,li2021brecq} and the large-model
methods it grew into~\cite{frantar2022obc,frantar2023gptq,lin2024awq,shao2024omniquant} optimise
reconstruction or output error and report language-model perplexity, as do the outlier-handling
schemes developed alongside them~\cite{dettmers2022llmint8,xiao2023smoothquant}; the question of when the calibration itself becomes harmful
is rarely asked because the loss is bounded by construction. Mixed-precision allocation by
Hessian sensitivity~\cite{dong2019hawq,dong2020hawqv2}, search~\cite{wang2019haq}, information
flow~\cite{akbulut2026infoq}, layer-wise geometry~\cite{lieq2025} or inter-layer
interaction~\cite{zhao2026coopq,mixquant2026} concerns \emph{which} modules to protect, an axis we
hold fixed. Dense
retrievers~\cite{karpukhin2020dpr,izacard2022contriever} are commonly evaluated on zero-shot suites such as
BEIR and MTEB~\cite{thakur2021beir,muennighoff2023mteb}, and retrieval-specific compression has
focused on the index and the output vectors~\cite{zhan2021jpq,zhan2022repconc}; our object is the encoder's weights and the failure mode
is a calibration dose, not a codebook. The closest methodological relative is early stopping with a
validation split, whose failure mode---flat validation curves that let an argmax wander---is exactly
what forced the interval rule of \S\ref{sec:gating}. Declining to calibrate when the evidence is
weak is also the stance of selective prediction~\cite{geifman2017selective}, applied here to a
deployment decision rather than to individual predictions.

\section{Conclusion}

Learned PTQ calibration of text embedders has a dose-driven unsafe regime that label-reading and
label-free objectives share. Inside the safe regime, ranking supervision can be genuinely useful
on some models and is not established as useful on others. A pre-registered, label-free validation
rule refused every harmful update on six independent calibration slates while forfeiting some
beneficial ones; the honest description is a safety--opportunity trade-off whose harm side is
observed at zero and whose opportunity side is measured at \NOppAcc/\NOppN. Calibration should be
treated as an optional intervention with a gate, not as an unconditional optimization.

\paragraph{Reproducibility.} The runner, the frozen protocol, the pre-registered comparator, the
selection rule, the audit log and the ledger from which every number in this paper is generated
are released with the paper.

\bibliographystyle{plain}
\bibliography{references}

\begin{thebibliography}{10}

\bibitem{akbulut2026infoq}
Mehmet~Emre Akbulut, Hazem Hesham~Yousef Shalby, Fabrizio Pittorino, and Manuel
  Roveri.
\newblock {InfoQ}: Mixed-precision quantization via global information flow.
\newblock In {\em Proceedings of the AAAI Conference on Artificial
  Intelligence}, 2026.
\newblock Quantizes each layer at each bit width and measures the resulting
  change in mutual information in the subsequent layers; allocates by integer
  linear programming. Evaluated on MobileNetV2 and ResNet18 / ImageNet.

\bibitem{bengio2013ste}
Yoshua Bengio, Nicholas L{\'e}onard, and Aaron Courville.
\newblock Estimating or propagating gradients through stochastic neurons for
  conditional computation.
\newblock {\em arXiv preprint arXiv:1308.3432}, 2013.

\bibitem{burges2007lambdarank}
Christopher J.~C. Burges, Robert Ragno, and Quoc~V. Le.
\newblock Learning to rank with nonsmooth cost functions.
\newblock In {\em Advances in Neural Information Processing Systems (NeurIPS)},
  2007.

\bibitem{chen2024bgem3}
Jianlv Chen, Shitao Xiao, Peitian Zhang, Kun Luo, Defu Lian, and Zheng Liu.
\newblock {M3}-embedding: Multi-linguality, multi-functionality,
  multi-granularity text embeddings through self-knowledge distillation.
\newblock In {\em Findings of the Association for Computational Linguistics:
  ACL 2024}, 2024.

\bibitem{choi2018pact}
Jungwook Choi, Zhuo Wang, Swagath Venkataramani, Pierce I-Jen Chuang,
  Vijayalakshmi Srinivasan, and Kailash Gopalakrishnan.
\newblock {PACT}: Parameterized clipping activation for quantized neural
  networks.
\newblock {\em arXiv preprint arXiv:1805.06085}, 2018.

\bibitem{cohan2020specter}
Arman Cohan, Sergey Feldman, Iz~Beltagy, Doug Downey, and Daniel~S. Weld.
\newblock {SPECTER}: Document-level representation learning using
  citation-informed transformers.
\newblock In {\em Proceedings of the Annual Meeting of the Association for
  Computational Linguistics (ACL)}, 2020.

\bibitem{conneau2020xlmr}
Alexis Conneau, Kartikay Khandelwal, Naman Goyal, Vishrav Chaudhary, Guillaume
  Wenzek, Francisco Guzm{\'a}n, Edouard Grave, Myle Ott, Luke Zettlemoyer, and
  Veselin Stoyanov.
\newblock Unsupervised cross-lingual representation learning at scale.
\newblock In {\em Proceedings of the Annual Meeting of the Association for
  Computational Linguistics (ACL)}, 2020.

\bibitem{dettmers2022llmint8}
Tim Dettmers, Mike Lewis, Younes Belkada, and Luke Zettlemoyer.
\newblock {LLM.int8()}: 8-bit matrix multiplication for transformers at scale.
\newblock In {\em Advances in Neural Information Processing Systems (NeurIPS)},
  2022.

\bibitem{devlin2019bert}
Jacob Devlin, Ming-Wei Chang, Kenton Lee, and Kristina Toutanova.
\newblock {BERT}: Pre-training of deep bidirectional transformers for language
  understanding.
\newblock In {\em Proceedings of the Conference of the North American Chapter
  of the Association for Computational Linguistics (NAACL)}, 2019.

\bibitem{diggelmann2020climatefever}
Thomas Diggelmann, Jordan Boyd-Graber, Jannis Bulian, Massimiliano Ciaramita,
  and Markus Leippold.
\newblock {CLIMATE-FEVER}: A dataset for verification of real-world climate
  claims.
\newblock {\em arXiv preprint arXiv:2012.00614}, 2020.

\bibitem{dong2020hawqv2}
Zhen Dong, Zhewei Yao, Daiyaan Arfeen, Amir Gholami, Michael~W. Mahoney, and
  Kurt Keutzer.
\newblock {HAWQ-V2}: Hessian aware trace-weighted quantization of neural
  networks.
\newblock In {\em Advances in Neural Information Processing Systems (NeurIPS)},
  2020.

\bibitem{dong2019hawq}
Zhen Dong, Zhewei Yao, Amir Gholami, Michael~W. Mahoney, and Kurt Keutzer.
\newblock {HAWQ}: Hessian aware quantization of neural networks with
  mixed-precision.
\newblock In {\em IEEE/CVF International Conference on Computer Vision (ICCV)},
  2019.

\bibitem{esser2020lsq}
Steven~K. Esser, Jeffrey~L. McKinstry, Deepika Bablani, Rathinakumar Appuswamy,
  and Dharmendra~S. Modha.
\newblock Learned step size quantization.
\newblock In {\em International Conference on Learning Representations (ICLR)},
  2020.

\bibitem{frantar2023gptq}
Elias Frantar, Saleh Ashkboos, Torsten Hoefler, and Dan Alistarh.
\newblock {GPTQ}: Accurate post-training quantization for generative
  pre-trained transformers.
\newblock In {\em International Conference on Learning Representations (ICLR)},
  2023.

\bibitem{frantar2022obc}
Elias Frantar, Sidak~Pal Singh, and Dan Alistarh.
\newblock Optimal brain compression: A framework for accurate post-training
  quantization and pruning.
\newblock In {\em Advances in Neural Information Processing Systems (NeurIPS)},
  2022.

\bibitem{geifman2017selective}
Yonatan Geifman and Ran El-Yaniv.
\newblock Selective classification for deep neural networks.
\newblock In {\em Advances in Neural Information Processing Systems (NeurIPS)},
  2017.

\bibitem{gemma2025gemma3}
{Gemma Team}, Aishwarya Kamath, Johan Ferret, Shreya Pathak, et~al.
\newblock {Gemma 3} technical report.
\newblock {\em arXiv preprint arXiv:2503.19786}, 2025.

\bibitem{gholami2021survey}
Amir Gholami, Sehoon Kim, Zhen Dong, Zhewei Yao, Michael~W. Mahoney, and Kurt
  Keutzer.
\newblock A survey of quantization methods for efficient neural network
  inference.
\newblock {\em arXiv preprint arXiv:2103.13630}, 2021.

\bibitem{izacard2022contriever}
Gautier Izacard, Mathilde Caron, Lucas Hosseini, Sebastian Riedel, Piotr
  Bojanowski, Armand Joulin, and Edouard Grave.
\newblock Unsupervised dense information retrieval with contrastive learning.
\newblock {\em Transactions on Machine Learning Research (TMLR)}, 2022.

\bibitem{karpukhin2020dpr}
Vladimir Karpukhin, Barlas O{\u{g}}uz, Sewon Min, Patrick Lewis, Ledell Wu,
  Sergey Edunov, Danqi Chen, and Wen-tau Yih.
\newblock Dense passage retrieval for open-domain question answering.
\newblock In {\em Proceedings of the Conference on Empirical Methods in Natural
  Language Processing (EMNLP)}, 2020.

\bibitem{kingma2015adam}
Diederik~P. Kingma and Jimmy Ba.
\newblock Adam: A method for stochastic optimization.
\newblock In {\em International Conference on Learning Representations (ICLR)},
  2015.

\bibitem{li2021brecq}
Yuhang Li, Ruihao Gong, Xu~Tan, Yang Yang, Peng Hu, Qi~Zhang, Fengwei Yu, Wei
  Wang, and Shi Gu.
\newblock {BRECQ}: Pushing the limit of post-training quantization by block
  reconstruction.
\newblock In {\em International Conference on Learning Representations (ICLR)},
  2021.

\bibitem{lin2024awq}
Ji~Lin, Jiaming Tang, Haotian Tang, Shang Yang, Xingyu Dang, Chuang Gan, and
  Song Han.
\newblock {AWQ}: Activation-aware weight quantization for on-device {LLM}
  compression and acceleration.
\newblock In {\em Proceedings of Machine Learning and Systems (MLSys)}, 2024.

\bibitem{mixquant2026}
Ashitabh Misra, Madhav Agrawal, Arham Jain, and Tarek Abdelzaher.
\newblock {MixQuant}: Adaptive mixed-precision quantization for large language
  models.
\newblock {\em arXiv preprint arXiv:2607.23047}, 2026.
\newblock Argues that a layer's sensitivity depends on the bit widths assigned
  to its upstream layers, and marginalizes each layer's distortion over
  randomly quantized upstream configurations to obtain budget-agnostic scores.

\bibitem{muennighoff2023mteb}
Niklas Muennighoff, Nouamane Tazi, Lo{\"i}c Magne, and Nils Reimers.
\newblock {MTEB}: Massive text embedding benchmark.
\newblock In {\em Proceedings of the 17th Conference of the European Chapter of
  the Association for Computational Linguistics (EACL)}, 2023.

\bibitem{nagel2020adaround}
Markus Nagel, Rana~Ali Amjad, Mart van Baalen, Christos Louizos, and Tijmen
  Blankevoort.
\newblock Up or down? adaptive rounding for post-training quantization.
\newblock In {\em International Conference on Machine Learning (ICML)}, 2020.

\bibitem{nguyen2016msmarco}
Tri Nguyen, Mir Rosenberg, Xia Song, Jianfeng Gao, Saurabh Tiwary, Rangan
  Majumder, and Li~Deng.
\newblock {MS MARCO}: A human generated {MAchine} reading {COmprehension}
  dataset.
\newblock In {\em CoCo@NeurIPS}, 2016.

\bibitem{shao2024omniquant}
Wenqi Shao, Mengzhao Chen, Zhaoyang Zhang, Peng Xu, Lirui Zhao, Zhiqian Li,
  et~al.
\newblock {OmniQuant}: Omnidirectionally calibrated quantization for large
  language models.
\newblock In {\em International Conference on Learning Representations (ICLR)},
  2024.

\bibitem{thakur2021beir}
Nandan Thakur, Nils Reimers, Andreas R{\"u}ckl{\'e}, Abhishek Srivastava, and
  Iryna Gurevych.
\newblock {BEIR}: A heterogeneous benchmark for zero-shot evaluation of
  information retrieval models.
\newblock In {\em NeurIPS Datasets and Benchmarks}, 2021.

\bibitem{vera2025embeddinggemma}
Henrique~Schechter Vera, Sahil Dua, Biao Zhang, et~al.
\newblock {EmbeddingGemma}: Powerful and lightweight text representations.
\newblock {\em arXiv preprint arXiv:2509.20354}, 2025.

\bibitem{voorhees2020treccovid}
Ellen Voorhees, Tasmeer Alam, Steven Bedrick, Dina Demner-Fushman, et~al.
\newblock {TREC-COVID}: Constructing a pandemic information retrieval test
  collection.
\newblock {\em arXiv preprint arXiv:2005.04474}, 2020.

\bibitem{wadden2020scifact}
David Wadden, Shanchuan Lin, Kyle Lo, Lucy~Lu Wang, Madeleine van Zuylen, Arman
  Cohan, and Hannaneh Hajishirzi.
\newblock Fact or fiction: Verifying scientific claims.
\newblock In {\em Proceedings of the Conference on Empirical Methods in Natural
  Language Processing (EMNLP)}, 2020.

\bibitem{wang2019haq}
Kuan Wang, Zhijian Liu, Yujun Lin, Ji~Lin, and Song Han.
\newblock {HAQ}: Hardware-aware automated quantization with mixed precision.
\newblock In {\em IEEE/CVF Conference on Computer Vision and Pattern
  Recognition (CVPR)}, 2019.

\bibitem{wang2022e5}
Liang Wang, Nan Yang, Xiaolong Huang, Binxing Jiao, Linjun Yang, Daxin Jiang,
  Rangan Majumder, and Furu Wei.
\newblock Text embeddings by weakly-supervised contrastive pre-training.
\newblock {\em arXiv preprint arXiv:2212.03533}, 2022.

\bibitem{xiao2023smoothquant}
Guangxuan Xiao, Ji~Lin, Mickael Seznec, Hao Wu, Julien Demouth, and Song Han.
\newblock {SmoothQuant}: Accurate and efficient post-training quantization for
  large language models.
\newblock In {\em International Conference on Machine Learning (ICML)}, 2023.

\bibitem{lieq2025}
He~Xiao, Qingyao Yang, Dirui Xie, Wendong Xu, Zunhai Su, Runming Yang, Wenyong
  Zhou, Haobo Liu, Zhengwu Liu, and Ngai Wong.
\newblock Exploring layer-wise information effectiveness for post-training
  quantization in small language models.
\newblock {\em arXiv preprint arXiv:2508.03332}, 2025.
\newblock LieQ. The deployed score is a geometry-only diagnostic:
  spectral-entropy compactness of a projection's output measured against an
  untrained weight matrix of the same initialization, read as training-induced
  energy concentration. Drives per-layer bit allocation under an average-bit
  budget with no gradient updates and no inference-time perplexity probing.

\bibitem{zhan2021jpq}
Jingtao Zhan, Jiaxin Mao, Yiqun Liu, Jiafeng Guo, Min Zhang, and Shaoping Ma.
\newblock Jointly optimizing query encoder and product quantization to improve
  retrieval performance.
\newblock In {\em Proceedings of the 30th ACM International Conference on
  Information and Knowledge Management (CIKM)}, 2021.

\bibitem{zhan2022repconc}
Jingtao Zhan, Jiaxin Mao, Yiqun Liu, Jiafeng Guo, Min Zhang, and Shaoping Ma.
\newblock Learning discrete representations via constrained clustering for
  effective and efficient dense retrieval.
\newblock In {\em Proceedings of the 15th ACM International Conference on Web
  Search and Data Mining (WSDM)}, 2022.

\bibitem{zhao2026coopq}
Junchen Zhao, Ali Derakhshan, Jayden Hyman, Junhao Dong, Sangeetha Abdu~Jyothi,
  and Ian Harris.
\newblock {CoopQ}: Cooperative game inspired layerwise mixed precision
  quantization for {LLM}s.
\newblock In {\em Findings of the Association for Computational Linguistics
  (ACL)}, 2026.
\newblock Frames allocation as a cooperative game among layers; Shapley-based
  Progressive Quantization Estimation (SPQE) estimates layer sensitivities and
  inter-layer interactions, solved as a binary quadratic program over 2/4-bit
  assignments.

\end{thebibliography}

\end{document}